%% file: Main.tex
\documentclass{article}
\usepackage{spconf,amsmath,graphicx}
\usepackage{subfiles}
\usepackage{times}
\usepackage{epsfig}
\usepackage{amssymb}
\usepackage{subfiles}
\usepackage{color}
\usepackage{multirow}
\usepackage{enumitem}
\usepackage{makecell}

\usepackage{caption} 
\title{CS-VQA: Visual Question Answering with Compressively Sensed Images}
\name{Li-Chi Huang,$^{2}$ Kuldeep Kulkarni,$^{3}$ Anik Jha,$^{2}$ Suhas Lohit,$^{2}$ Suren Jayasuriya,$^{1,2}$ Pavan Turaga$^{1,2}$\thanks{This work was supported in part by ARO grant W911NF-17-1-0293.}}

\address{$^{1}$School of Arts, Media and Engineering, Arizona State University\\
$^{2}$School of Electrical, Computer and Energy Engineering, Arizona State University\\ 
$^{3}$Department of Electrical and Computer Engineering, Carnegie Mellon University}

\begin{document}
%
\maketitle
\begin{abstract}
Visual Question Answering (VQA) is a complex semantic task requiring both natural language processing and visual recognition. In this paper, we explore whether VQA is solvable when images are captured in a sub-Nyquist compressive paradigm. We develop a series of deep-network architectures that exploit available compressive data to increasing degrees of accuracy, and show that VQA is indeed solvable in the compressed domain. Our results show that there is nominal degradation in VQA performance when using compressive measurements, but that accuracy can be recovered when VQA pipelines are used in conjunction with state-of-the-art deep neural networks for CS reconstruction. The results presented yield important implications for resource-constrained VQA applications.  
\end{abstract}
\section{Introduction}
\label{sec:intro}
\subfile{intro.tex}

\section{Background and Related Work}
\label{sec:background}
\subfile{background.tex}

\section{Problem Formulation and Approach}
\label{sec:formulation}
\subfile{formulation.tex}

\section{Experiments}
\label{sec:experiments}
\subfile{experiments.tex}
\vspace{-0.4in}

\section{Conclusion}
\label{sec:analysis}
\subfile{analysis.tex}


\bibliographystyle{IEEEbib}
\begin{small}
\bibliography{refs}
\end{small}
\end{document}

%% file: intro.tex
The Visual Question Answering (VQA) problem has recently gained significant research attention in the computer vision and machine learning communities~\cite{VQA}. The VQA task consists of answering an open-ended question for a given image, which requires the ability to parse a question expressed in natural language, computationally analyze the image based on the question's requirement, and present an answer in natural language. For example, given an image depicting a family reunion, representative questions might include ``How many people are there?", ``What is the color of the table?" etc. Due to the contextual analysis required to answer these questions, VQA has been considered an AI complete task~\cite{VQA}. Contemporary VQA research has utilized deep neural-networks trained jointly on images and natural language `vectors' computed from the questions. However, in this paper, we explore whether the underlying representation of visual data in 2D images is even critical for VQA performance. In particular, we explore whether sub-Nyquist rate sensed measurements of natural images can be an effective substitute for fully-sampled images in a VQA architecture. 

The answer to the above question can have significant implications for adapting VQA techniques to resource-constrained platforms, such as a Google Glass, a Hololens, mobile computing platforms, field robotics etc. For instance, the Google Glass continuously running an off-the-shelf face-detection algorithm drains its battery in only $45$ minutes~\cite{likamwa2014draining}. Sub-Nyquist imagers hold the potential to save imaging energy, reducing data-bandwidth, storage, etc. all of which can result in sustaining performance under resource constraints.

The most popular sub-Nyquist, or, compressive sensing (CS) framework for imaging has utilized a sampling framework where incoming light-rays are multiplexed onto a smaller set of pixels (even a single pixel~\cite{duarte2008single}). Via multiple coded projections of a scene, the original image can be reconstructed using post-processing ~\cite{donoho2006compressed,candes2006robust}. This allows CS techniques to satisfy resource constraints in real imaging systems including decreasing energy consumption, computation, bandwidth, and latency.  Working with CS data requires rethinking the computer vision pipeline, as even basic operations like convolutions require non-trivial computation such as smashed filtering \cite{DavenportDWLTKB2007,kulkarni2016reconstruction}. We term this new task CS-VQA and present new approaches to solve this task.
\indent \textbf{Contributions:} This paper is a first investigation of the CS-VQA task. We design a series of deep neural-network architectures to solve CS-VQA. While some of the proposed modules are inspired from past work in CS reconstruction, we do not require explicit reconstruction. We also investigate whether CS imaging is more suited for answering certain types of questions, more so than others. Finally, we explore the tradeoffs between performance, computational time, size of models, etc., and show that it is indeed possible to achieve near state-of-the-art VQA performance, even while working with compressively sensed imagery.

%% file: background.tex
\indent \textbf{Compressive Sensing:}
Compressive Sensing (CS) is a signal acquisition paradigm which samples
a signal at sub-Nyquist rates using random linear measurements, and then recovers the original signal in post-processing~\cite{donoho2006compressed,candes2006robust}. The measurements are given by $\mathbf{y} = \Phi \mathbf{x} + \mathbf{e}$ , with image $\mathbf{x} \in \mathbb{R}^n$, measurement vector $\mathbf{y} \in \mathbb{R}^m$, measurement/projection matrix $\Phi \in \mathbb{R}^{m\times n}$ and additive noise $\mathbf{e} \in \mathbb{R}^m$. To solve this ill-posed problem when $m << n$, one can solve the following optimization problem in equation~\ref{eq:CS}, provided the signal is $s$-sparse in some sparsifying domain, $\Psi$, 
\begin{equation} 
\min_{\mathbf{x}} ||\Psi \mathbf{x}||_1 \text{ s.t. } ||\mathbf{y} - \Phi \mathbf{x}||_2 \leq \epsilon. 
\label{eq:CS}
\end{equation}

To solve \eqref{eq:CS}, many iterative algorithms have been proposed in the literature~\cite{candes2006near,donoho2006compressed, baraniuk2010model,kim2010compressed,donoho2009message} but they are not conducive for fast reconstruction or low measurement rates (MRs). For faster reconstruction and better recovery at low MRs ($< 0.10$), deep learning networks have been proposed that achieve state-of-the-art performance~\cite{mousavi2015deep,iliadis2018deep,kulkarni2016reconnet}.

\indent \textbf{Compressive Inference:} The goal of compressive inference is to infer semantic information directly from compressed measurements without reconstruction. Direct inference has been shown to be feasible in applications like action recognition~\cite{kulkarni2016reconstruction}, image classification~\cite{lohit2016direct}, and object tracking~\cite{kulkarni2016fast}. This paper explores the tradeoffs in compressive VQA, which has not been attempted in the past.

\indent \textbf{Visual Question Answering:} Current approaches to solve VQA rely heavily on deep-learning methods, for fusing image and text features. Image features are typically extracted using pre-trained or fine-tuned convolutional neural networks (CNNs) such as GoogleNet~\cite{szegedy2015going} or ResNet~\cite{he2016deep}. Textual questions are converted into vector sequences using methods such as Word2vec~\cite{mikolov2013distributed}, and further processed using recurrent neural networks (RNNs) or Long Short-Term Memory cells (LSTMs)~\cite{hochreiter1997long} to encode temporal structure within the questions. A late-stage fusion is done for the image and question features, and a final classifier provides an answer from a set of specified answers. Recent approaches have utilized attention mechanisms to spatially localize image features for improved performance~\cite{xu2016ask, yang2016stacked, lu2016hierarchical}. State-of-the-art VQA models use an ensemble of methods~\cite{fukui2016multimodal}. In this paper, we do not seek to improve VQA performance, but  investigate the effect of sub-Nyquist sensing of images on VQA performance.

%% file: formulation.tex
\begin{figure*}[h!]
\centering
\includegraphics[width=\textwidth]{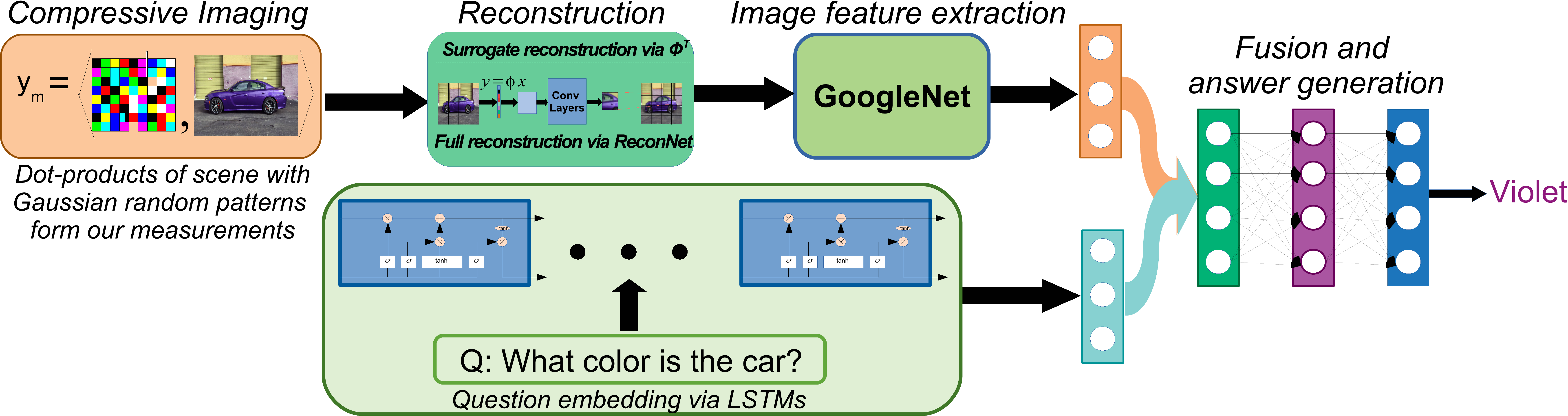}
\caption{An overview of the proposed CS-VQA architecture. An image is compressively sensed via random projections. The measurement vector is used to reconstruct either a surrogate image using linear inversion or full-reconstruction via ReconNet~\cite{kulkarni2016reconnet}. The reconstructed image is used to extract visual features using GoogleNet~\cite{szegedy2015going}. The given question is encoded using Word2vec~\cite{mikolov2013distributed} and fed into a LSTM to form a question feature vector. The two features are concatenated and fed into a fully connected network to generate the final answer. }
\label{fig:sys}
\vspace{-0.2in}
\end{figure*}

Our goal is to answer questions posed with respect to a scene, given its CS measurements. The existing VQA dataset~\cite{VQA} consists of image-question pairs, thus we must convert the images into measurement vectors. We simulate compressive sensing with either a random Gaussian or column-permuted Hadamard measurement matrix, operating at a measurement rate (MR) = $m/n$. The choice of the measurement matrix is motivated by the following reasons, a) it is task-agnostic, yet generalizable to many tasks, b) it is a theoretically supported method for compressive acquisition of natural image data. However, learning a measurement matrix may result in improved performance, but we leave this avenue for future work. For our experiments, we simulate compressive sensing two different ways, each yielding a CS-VQA dataset, and conduct experiments on both the CS-VQA datasets. 

In the first method, the measurements $\mathbf{y}$ are obtained by pre-multiplying the image vector, $\mathbf{x} \in \mathbb{R}^n$ by a column-permuted Hadamard matrix ($\Phi$) of size $m \times n$, mathematically written as $\mathbf{y} = \Phi \mathbf{x}$. We call this `FF-CS-VQA' (FF $=$ full-frame). In the second, the image is divided into non-overlapping blocks of a fixed size, $33 \times 33$, and independent measurements are obtained for each block, using a common random Gaussian measurement matrix, $\mathbf{y}_B = \Phi_B \mathbf{x}_B$. We call this as `B-CS-VQA' (B $=$ block).

\indent \textbf{Reconstruction and Network Architecture:}
Most common VQA architectures consists of two streams -- one which operates on the given image and outputs a visual feature, and the other which operates on a word-embedding of the question and outputs a text feature. These feature vectors are concatenated, and further processed by a small network of fully-connected layers to obtain probability scores over the set of possible answers to the question. However, as described above, our CS-VQA dataset consists of modulated CS measurements. Hence, we need to redesign the image feature stream. We use four different approaches to recover surrogates of the image from its compressive measurement, each with different levels of sophistication. An overview of our CS-VQA architecture is shown in Figure~\ref{fig:sys}. We investigate the following CS surrogate-reconstruction approaches. 
\begin{itemize}[leftmargin=*]
\item Raw Multiplexed: This is when we \textbf{do not} perform any CS reconstruction, but use the raw CS measurement vectors as image features directly (i.e. no need for visual feature extraction using GoogleNet). 

\item $\Phi^T \mathbf{y}$: For each image, $I$, in the FF-CS-VQA dataset, we apply the transformation $\Phi^T$ to the measurement vector, $\mathbf{y}$ to obtain $\Phi^T \mathbf{y}$ which is reshaped to the image size. 

\item Block-wise linear inversion, $\Phi_B^T \mathbf{y}_B$: In the B-CS-VQA dataset, we apply the transformation $\Phi_B^T$ to the measurement vectors for each non-overlapping block, and reshape the transformed vectors to the size of the image block. The reshaped blocks are arranged on a 2D grid, given by $I$.

\item ReconNet: For each image, $I$, in the B-CS-VQA dataset, we use ReconNet \cite{kulkarni2016reconnet}, to obtain the reconstructed images. This corresponds to full reconstruction.
\end{itemize}

\indent \textbf{Visual Feature Extraction:} After CS surrogate reconstruction, we use GoogleNet~\cite{szegedy2015going} to extract visual features. To train GoogleNet on these surrogates, we employ the following scheme: (1) initialize with pre-trained weights from the ImageNet dataset~\cite{russakovsky2015imagenet}, and then (2) fine-tune the network by performing image classification on CS surrogate-reconstructions. Given an image $I$, we obtain a 1024-length feature representation for the image by tapping the output of the penultimate layer of the GoogleNet, denoted by $\mathbf{v}_I$.
\indent \textbf{Question Embedding:}
Questions are encoded using Word2vec \cite{mikolov2013distributed}, such that the input to the LSTM is a sequence $Q = \left( \mathbf{w}_1, \ldots, \mathbf{w}_N \right)$. We employ an LSTM that is identical to that of~\cite{VQA}. The LSTM states represent sequence embeddings, $\mathbf{h}_t = LSTM(\mathbf{w}_t,\mathbf{h}_{t-1})$, $\mathbf{h}_0$ is an all-zero vector. The question embedding is the final state of the LSTM $\mathbf{q}_I=\mathbf{h}_T$~\cite{VQA}.
\indent \textbf{Fusing Visual and Language Features:} We use simple concatenation to fuse image and question feature vectors. This fused  vector is fed into a fully connected network. 

%% file: experiments.tex
In this section, we evaluate the proposed architectures for the proposed CS-VQA task. The VQA dataset~\cite{VQA} uses images from the MS COCO dataset~\cite{coco}, which contains $83783$ training images and $40504$ validation images. The dataset includes three questions for each image, so there are a total of $248349$ questions for the training set and $121512$ questions for the validation set. Answers for questions are generated by Amazon Mechanical Turk (AMT) annotators, with 10 answers per question from unique annotators. Answers are generally open-ended, types of answers are generally classified as  ``yes and no'', ``number" and ``other" answers. We adopt the validation set to test the performance of the proposed approach. The evaluation metric for the open-ended task in VQA dataset given a generated answer is as following: 
\begin{equation}\label{eq:vqaac}
\text{Accuracy} = \min\left(\frac{\text{\# of matches to ground truth}}{3},1\right).
\end{equation}
This metric gives the answer full credit if the generated answer matches with at least three (of ten) answers provided by AMT annotators. Otherwise, it is given partial credit. 

\indent \textbf{Training Details:} GoogLeNet was finetuned on Caffe, whereas TensorFlow framework was used to train and test the LSTM unit. All training and testing was performed on an NVIDIA Titan X GPU. For finetuning GoogleNet a batch size of $32$ images was used, with data augmentation by mirror reflection of images. At MR = 0.25, stochastic gradient descent (SGD) was used with momentum 0.9, initial learning rate of 0.001 and learning rate decay of 0.8 for every 80000 iterations. A dropout of 0.4 was used on the last fully connected layer. For LSTM training, Adam optimizer was used with initial learning rate of 0.0003 and learning rate decay of 0.999 for every 5000 iterations. A dropout of 0.5 was used on each LSTM layer. Finetuning takes about 7 days when starting from pre-trained GoogleNet. 


\begin{table}[h]

\begin{small}
\begin{tabular}{|c|c|c|c|c|}
\hline
\multirow{2}{*}{\makecell{CS Reconstruction\\Method}}  & \multicolumn{4}{c|}{Question Type} \\
\cline{2-5}
& All & Yes/No & Number & Other   \\ \hline \hline
None (Raw Multiplexed)   & 47.95 & 78.34 & 32.45 & 29.10   \\ \hline
$\Phi^T \Phi \mathbf{x}$    &    51.10  & 78.82  & 33.30  & 34.82 \\ \hline
$\Phi_B^T \Phi_B \mathbf{x}_B$     &      52.98                 &     79.50                  &     33.03                  &        38.15      \\ \hline

ReconNet   &        54.22               &   79.85                    &      33.28                 &         40.21     \\ \hline 
\hline
\multicolumn{5}{|c|}{Oracle VQA~\cite{VQA}}                                                                                                                                               \\ \hline \hline
LSTM + VGG                                             &        57.75               &   80.50                    &       36.77                &          43.08             \\ \hline
Image Only        & 28.13 & 64.01 & 0.42 & 3.77 \\ \hline
Question Only  & 50.39 & 78.41 & 34.68 & 30.03 \\ \hline
\end{tabular}%

\end{small}
\caption{Open-ended VQA v1.0 results with various CS surrogate-reconstructions, and their corresponding accuracy (\%). GoogleNet was used for visual feature extraction, and a LSTM for generating question features. The oracle VQA~\cite{VQA} performance is presented for comparison. }
 \label{tb:main}
 \vspace{-0.1in}
 \end{table}



\textbf{Main Results:} In Table~\ref{tb:main}, we show the results of open-ended VQA performance for various different CS reconstruction techniques at MR = 0.25. We compare this to the original results from the VQA paper~\cite{VQA}, which we term the Oracle VQA. Note that training directly on CS measurements themselves (Raw Multiplexed) yields a 10\% point drop in performance, and is mostly comparable to the question-only baseline (i.e. when no visual information is used). Each reconstruction technique, $\Phi^T\Phi, \Phi^T_B\Phi_B,$ ReconNet yields improvement to their performance, particularly in the ``other" question category. Note that this question category seems to rely the most on visual data as evidenced by the Oracle VQA performance presented. ReconNet performs the best of the proposed methods, and is within 3\% points of the oracle VQA.

\begin{figure}
\centering
\includegraphics[width = 0.48\textwidth]{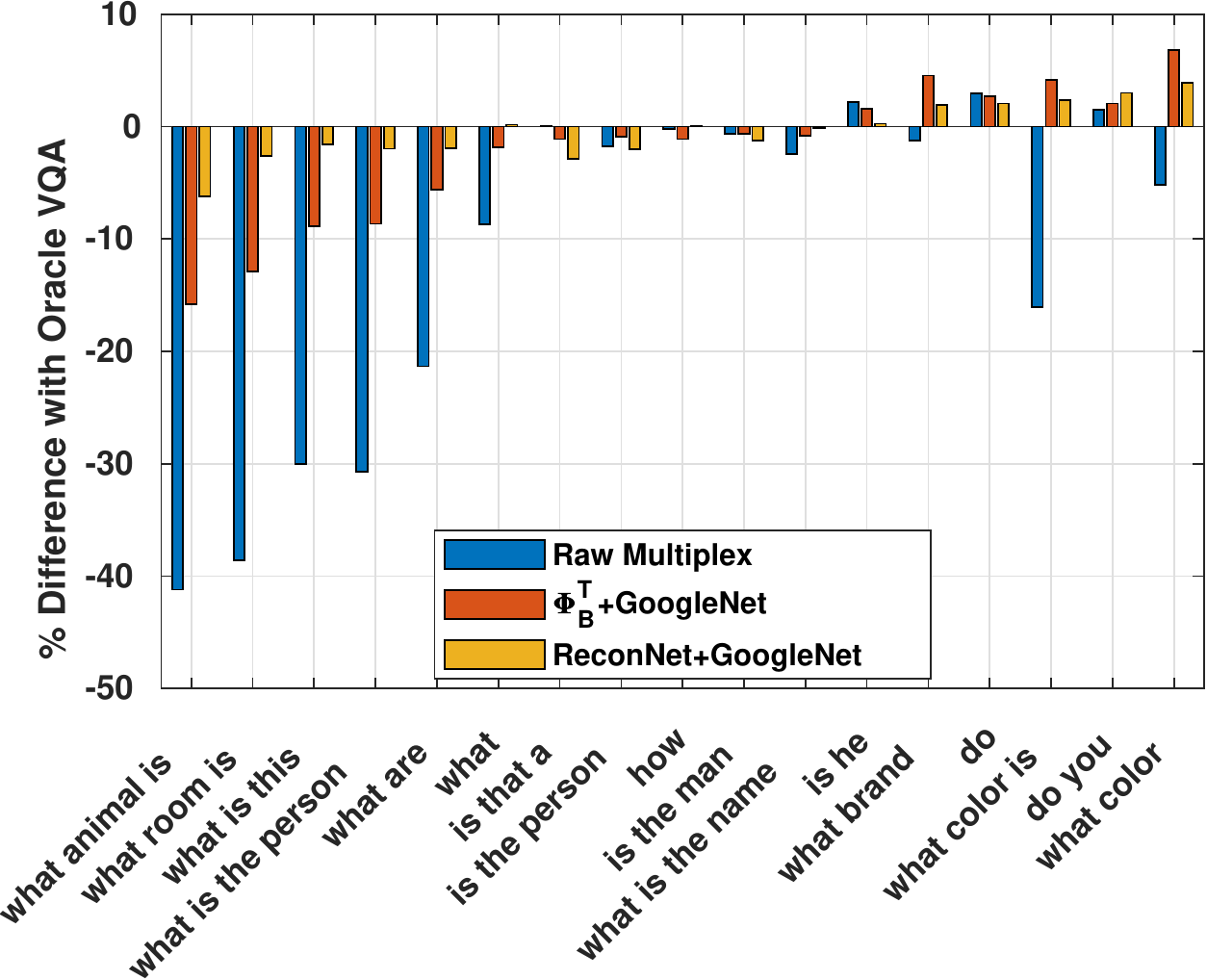}
\caption{Comparison of CS-VQA to the Oracle VQA~\cite{VQA} for different types of questions.}
\label{fig:compare}
\vspace{-0.2in}
\end{figure}

In Figure~\ref{fig:compare}, we show the results of three different models: Raw Multiplexed, $\Phi^T_B$, and ReconNet with respect to the oracle VQA algorithm, sorted by question category. The questions where the oracle method outperforms the three CS-VQA methods, typically feature a \textit{specific} question about a subject/object in the picture, including ``what animal is", ``what room is", ``what is the person", ``what sport is". In contrast, questions such as ``what color", ``is there", ``do you" are better answered by the CS algorithms. 

\indent \textbf{Measurement Rate:} We also tested the effect of varying the measurement rate on the results.  At MR = 0.10, ReconNet's VQA accuracy is 51.40\% with a breakdown of 79.13\% yes/no, 33.20\% number, and 35.21\% other. At MR = 0.01, ReconNet's VQA accuracy is 51.05\% with a breakdown of 78.77\% yes/no, 32.92\% number, and 34.87\% other. This validates that reconstructions at low measurement rates still perform well on the VQA task.

\indent \textbf{VQA v2.0:} We also compared the performances of $\Phi^T_B$, and ReconNet based CS reconstruction models on the open-ended questions of VQA v2.0 dataset with that of the Oracle-VQA~\cite{vqa_v2} and tabulated them in Table~\ref{tb:vqa2}. Their comparable performances indicate that CS-VQA is also able to effectively handle the reduced language bias in VQA v2.0 dataset.\newline

\begin{table}[h!]
\centering

\begin{small}
\begin{tabular}{|c|c|c|c|c|}
\hline
\multirow{2}{*}{\makecell{CS Reconstruction\\Method}}  & \multicolumn{4}{c|}{Question Type} \\
\cline{2-5}
& All & Yes/No & Number & Other   \\ \hline \hline
$\Phi_B^T \Phi_B \mathbf{x}_B$     &      48.92                 &     70.61                  &     33.13                  &        36.58      \\ \hline

ReconNet   &        49.85               &   70.50                    &      33.32                 &         38.52     \\ \hline 
\multicolumn{5}{|c|}{Oracle VQA~\cite{vqa_v2}}                                                                                                                                               \\ \hline \hline
LSTM + VGG                                             &        54.22               &   73.46                    &       35.18                &          41.83            \\ \hline
Question Only  & 44.26 & 67.01 & 31.55 & 27.37 \\ \hline
\end{tabular}%

\end{small}
\caption{Open-ended VQA v2.0 results with various CS reconstructions, and their corresponding accuracy(\%) }
 \label{tb:vqa2}
 \vspace{-0.1in}
 \end{table}

\textbf{Run-time Complexity of Models:} The average execution times for each model to answer a question, for one image, is presented in Table~\ref{tb:extime}. We average the results of the Caffe ``time" command over 5 runs. The command uses random weights for measuring the time, and each computation time obtained for each of the 5 runs is itself the average over 100 iterations of forward pass through the network. All the numbers except for $\Phi^T$ are obtained using Caffe on Titan X GPU, with $\Phi^T$ executed on a CPU with Matlab due to space considerations (too large to fit on the GPU). We can see from the table that all three methods are considerably faster than a traditional iterative CS solver, but ReconNet gives the best VQA performance with relatively fast execution time.   

\begin{table}
\begin{center}
\begin{tabular}{|c|c|}
\hline
Reconstruction Method & Time (ms)\\ \hline \hline
$\Phi^T \Phi \mathbf{x}$ &  29.36    \\   
\hline
Block-based $\Phi^T_B \Phi_B \mathbf{x}$   &  20.78   \\   
\hline
ReconNet &  27.99 \\ 
\hline
TVAL3 (from~\cite{kulkarni2016reconnet}) & 2963.00 \\
\hline
\end{tabular}
\end{center}
\caption{Average execution time per image to generate answers, for various models.}
\label{tb:extime}
\vspace{-0.1in}
\end{table}

In addition to execution speed benefits, we also compare the memory requirements in terms of the number of parameters in each model. Using $\Phi_B^T \Phi_B \mathbf{x}$ (at MR = 0.25) with GoogleNet and the LSTM (including the fusion layers) results in 12,610,768 parameters. Using ReconNet (at MR = 0.25) along with the same back-end of GoogleNet and LSTM results in 12,633,488 parameters. This is only a slight increase for an improvement of 1-2\% points on the CS-VQA task, and an extra $8$ms of processing time. Using raw multiplexed measurements requires only 6,644,496 parameters. 
\\
\\



%% file: analysis.tex
In summary, we have presented the first study of the effectiveness of VQA on compressively sensed images. In particular, we show that VQA can achieve near-equivalent performance to natural images when using advanced compressive sensing (CS) reconstruction techniques such as ReconNet with a performance gap of only 3\% points at measurement rate MR = 0.25, and 6\% gap at MR = 0.01. Using direct inference approaches, we report reduced processing time over approaches that need full reconstruction, and reduced network parameters. Of course, using a full-reconstruction approach results in the best performance. We believe this work opens up a new avenue of research into VQA for derived or intermediate representations of visual data which are amenable to system considerations such as energy-efficiency and limited bandwidth for mobile and embedded AI platforms.